\documentclass[journal]{IEEEtran}

\ifCLASSINFOpdf
\else
   \usepackage[dvips]{graphicx}
\fi
\usepackage{url}

\usepackage{array}
\usepackage{color}
\usepackage{amstext, amssymb, amsthm, amsmath,bm,rotating,stmaryrd}
\usepackage{booktabs,url}
\usepackage{subfig}
\usepackage{graphicx}
\usepackage{epstopdf}
\usepackage{algorithm}
\usepackage{algpseudocode}
\usepackage{multirow}
\usepackage[numbers]{natbib}

\newtheorem{theorem}{Theorem}

\newcommand{\vx}{{\bf x}}
\newcommand{\vs}{{\bf s}}

\newcommand{\vw}{{\bf w}}
\newcommand{\vy}{{\bf y}}

\newcommand{\valpha}{{\bf \alpha}}

\newcommand{\vC}{{\bf C}}

\newcommand{\vA}{{\bf A}}
\newcommand{\vV}{{\bf V}}

\newcommand{\vX}{{\bf X}}
\newcommand{\vW}{{\bf W}}
\newcommand{\vI}{{\bf I}}
\newcommand{\vK}{{\bf K}}

\begin{document}

\title{One-shot Distributed Algorithm for PCA with RBF Kernels}

\author{Fan He, Kexin Lv, Jie Yang, and Xiaolin Huang, \IEEEmembership{Senior Member, IEEE}
\thanks{This work is supported by National Natural Science Foundation of China (No. 61977046, 11631015, U1803261) and Program of Shanghai Subject Chief Scientist (18XD1400700).}
\thanks{F. He, K. Lv, J. Yang and X. Huang are with the Institute of Image Processing and Pattern Recognition, Shanghai Jiao Tong University, also with the MOE Key Laboratory of System Control and Information Processing, 800 Dongchuan Road, Shanghai, 200240, P. R. China. (e-mail: hf-inspire@sjtu.edu.cn, kelen\_lv@sjtu.edu.cn, jieyang@sjtu.edu.cn, xiaolinhuang@sjtu.edu.cn).}}

\markboth{ }
{He \MakeLowercase{\textit{et al.}}: One-shot Distributed Algorithm for PCA with RBF Kernels}
\maketitle

\begin{abstract}
This letter proposes a one-shot algorithm for feature-distributed kernel PCA.
Our algorithm is inspired by the dual relationship between sample-distributed and feature-distributed scenario.
This interesting relationship makes it possible to establish distributed kernel PCA for feature-distributed cases from ideas in distributed PCA in sample-distributed scenario. 
In theoretical part, we analyze the approximation error for both linear and RBF kernels.
The result suggests that when eigenvalues decay fast,
the proposed algorithm gives high quality results with low communication cost.
This result is also verified by numerical experiments, showing the effectiveness of our algorithm in practice.
\end{abstract}


\IEEEpeerreviewmaketitle

\section{Introduction}

\IEEEPARstart{P}{rincipal} Component Analysis (PCA) is a fundamental technology in machine learning. For nonlinear tasks, PCA could be formulated as the following problem,
\begin{equation}\label{equ: kpca}
    \begin{aligned}
        \max_{\vw^\top\vw = \vI} \|\vw^\top\phi(\vX)\|_F^2.
    \end{aligned}
\end{equation}
Here $\vX\in\mathcal{R}^{M\times T}$ denotes data and $\phi(\cdot):\mathcal{R}^{M}\rightarrow \mathcal{F}$ is a unknown non-linear mapping. Interestingly, the kernel trick could be implemented there, resulting the kernel PCA (KPCA)  \cite{mika1999kernel, scholkopf1998nonlinear, scholkopf1999kernel}.
Specifically, the optimal solution of (\ref{equ: kpca}) is $\vw = \phi(\vX)\valpha$, where $\valpha$ is the eigenvectors of the gram kernel matrix $\vK \triangleq \langle \phi(\vX), \phi(\vX)\rangle$.

\begin{figure}[tb]
    \centering
    \includegraphics[width=0.45\textwidth]{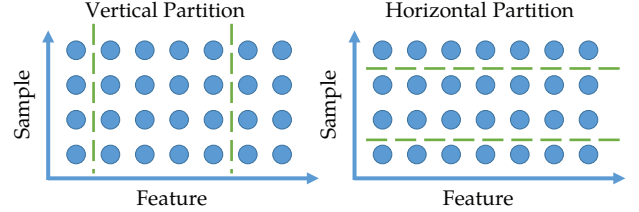}
    \caption{Categorizations of data partition in distributed setting.}
    \label{fig: partition}
\end{figure}

Nowadays, distributed algorithm of KPCA is in high demand. 
Generally, distributed data could be categorised as two regimes, namely horizontally and vertically partitioned data \cite{yang2019federated, fan2019distributed}. 
As shown in Fig.~\ref{fig: partition}, the horizontal and vertical axes are features and samples.
Then, when data are partitioned horizontally, agents contain part of samples with all features; when data are partitioned vertically, agents contain full samples but with only part of features.

For PCA problems, there are massive researches focusing on both the horizontal  \cite{fan2019distributed,ge2018minimax, garber2017communication} and the vertical regime \cite{scaglione2008decentralized,li2011distributed, SchizasA}. 
However, although KPCA are very common in practice, e.g., in face recognition \cite{kim2002face, liu2004gabor} and process monitoring \cite{choi2004nonlinear, liu2009moving}, applicable distributed algorithms are not much. 
To the best of our knowledge, the existing studies on distributed KPCA are all for the horizontal regime, most of which require multi-communication rounds; see, e.g., 
\cite{rosipal2001an}, \cite{zheng2005an}, \cite{balcan2016communication}. 

In this letter, we aim to develop distributed KPCA for vertically partitioned data. The main obstacle in vertical regime is that we only locally know a part of features, then it seems that one cannot know the global kernel without heavy communication. To handle this problem, we fully investigate the kernel trick that 
can transfer the primal optimization variables corresponding to features to dual variables corresponding to samples.
With this idea, we can 
link KPCA in the vertical regime to PCA in the horizontal regime 
and establish a distributed KPCA (DKPCA) in the vertical regime. 
The proposed DKPCA locally calculates the 
the first $D$ eigenvectors and their corresponding eigenvalues of local kernel matrices and then sent to a fusion center, where they are aggregated to reproduce local estimators.
DKPCA is applicable to both linear and RBF kernel and it needs only one  communication round.
Theoretical discussion will show that the approximation error is related to the $D$-th eigenvalue of local matrices. When eigenvalues decay fast, 
DKPCA has very high quality results, which is also confirmed by numerical experiments. 


\section{Distributed kernel PCA}\label{sec: alg}

\subsection{Preliminaries and Notations}
We use regular letters for scalars, capital letters in bold for matrices and lowercase letters in bold for vectors.
For a matrix $\vA$, $\|\vA\|_F$ represents the Frobenius norm.
We consider a distributed setting, where the data are partitioned vertically and stored distributedly in $J$ local agents.
Such vertical regime are common in practice, e.g., in wireless sensor networks \cite{ghorbel2015distributed,le2008distributed}, ranking or evaluation systems \cite{kargupta2001distributed, mcmahan2016communication}.
Specifically, each agent $j$ acquires a zero-mean data vector $\vx^{(j)} = \{\vx_i^{(j)}\}_{i=1}^T \in \mathcal{R}^{M_j\times T}$, which is i.i.d. at time $i = 1,2,\cdots, T$. $M_j$ is the feature dimension of the data $\vx^{(j)}$ and we have $\sum_{j=1}^J M_j = M$.
Let $\vX =[(\vx^{(1)})^\top (\vx^{(2)})^\top \cdots (\vx^{(J)})^\top]^\top \in \mathcal{R}^{M\times T}$ denote all data collected by agents, which are not stored together but given for convenience.

Here we briefly review the KPCA problem, i.e., problem (\ref{equ: kpca}),
where PCA is executed in a Reproducing Kernel Hilbert Space (RKHS) introduced by a reproducing kernel $\kappa(\cdot, \cdot)$.
The goal of KPCA is to diagonalize the covariance matrix $\vC=\sum_i\phi(\vx_i)\otimes\phi(\vx_i)^*$.
However, it is hard to do eigendecomposition on $\vC$ since $\phi$ is implicit.
Then KPCA turns to solve the dual eigenproblem, where the eigendecomposition is performed on the gram kernel matrix $\vK\in\mathcal{R}^{T\times T}$, i.e.,
\begin{equation*}
\begin{aligned}
    &\vK_{p,q} = \kappa(\vx_p,\vx_q) = \langle\phi(\vx_p), \phi(\vx_q)\rangle, p,q=1,\cdots,T,\\
	&\lambda_d {\valpha}_d = \vK{\valpha}_d, d=1,\cdots,T,
	\end{aligned}
\end{equation*}
where $\valpha_d\in\mathcal{R}^{T\times 1}$ is the eigenvector of $\vK$ corresponding to the $d$-th largest eigenvalue $\lambda_d$.
Then the $d$-th eigenvector of $\vC$ can be rewritten as 
$\vw_d = \phi(\vX)\valpha_d$.
Such kernel trick sidesteps the problem of computing unknown $\phi(\cdot)$ and moreover, it makes the distributed computation for vertically partitioned data more convenient because:
\begin{itemize}
    \item the covariance $\vC$ is not separable and generally the approximation by local covariance matrices is not accurate, e.g., $\vC \neq \sum_j \vC^{(j)} =\sum_j\sum_i \phi(\vx_i^{(j)})\otimes\phi(\vx_i^{(j)})^*$. 
    \item $\vK$ itself (linear kernels) or its main calculation part (RBF kernels) is separable, e.g., a linear kernel $\vK = \sum_j \vK^{(j)}$.
\end{itemize}

\subsection{Distributed algorithm for kernel PCA}

\begin{algorithm}[b]
    \begin{algorithmic} [1]
    \caption{One-shot distributed algorithm for kernel PCA in the vertical partition regime. (DKPCA)}
    \label{alg: DKPCA-os}
    \State On local agents, calculate the local kernel matrix $\vK^{(j)}$. Solve eigenvalue problem on $\vK^{(j)}$, obtaining $\vV_D^{(j)}$ and $\bm \lambda_D^{(j)}$ and sent them to the fusion center.    	
	\State On the fusion center, calculate $\hat{\vK}$ to approximate $\vK$.
	
	 - For linear kernel, use (\ref{equ: agg-linear}).
	 
     - For RBF kernel, use (\ref{equ: agg-rbf}).
    \State Compute the leading $D$ eigenvectors $\hat{\vV} \in \mathcal{R}^{T\times D}$ of the approximate matrix $\hat{\vK}$.    
    \State \Return $\hat{\vV}$.
    \end{algorithmic}
\end{algorithm}

We first link the existing distributed algorithm for PCA in horizontal regime \cite{fan2019distributed} to that for KPCA in vertical regime, and then extend it to RBF kernels. The proposed algorithm could produce a good estimation to the global optimum in one-communication round with privacy protection.

In the horizontal regime, the key property is the consistency between the sum of local covariance matrices and the global covariance matrix, i.e., $\vC=\sum_j\vC_j$, which results in benefits for both algorithm design and theory analysis.
Considering linear KPCA in vertical regime is a dual problem of linear PCA in horizontal regime, we can easily transfer the algorithm in \cite{fan2019distributed} to that of KPCA. Mathematically, for linear kernels, $\phi(\vx) = a\vx, \forall a\in\mathcal{R}$. We assume $a = 1$ without loss of generality and then it holds that
$\vK = \vX^\top \vX = \sum_{j = 1}^J (\vx^{(j)})^\top \vx^{(j)}= \sum_j \vK^{(j)}$.
Thus, the estimator $\hat{\vK}$ is calculated as follow.
\begin{equation}\label{equ: agg-linear}
	\hat{\vK} =\sum\nolimits_j \hat{\vK}^{(j)} = \sum\nolimits_j  \vV_D^{(j)}\bm{\lambda}_D^{(j)}(\vV_D^{(j)})^\top.
\end{equation}

However, in horizontal regime, algorithm in \cite{fan2019distributed} cannot be extend to non-linear case, limiting its application in practice.
Instead, in vertical regime, we can further extend this idea to RBF kernels by noting the following property.
\begin{equation}\label{equ: element RBF}
\begin{aligned}
	&\vK(p,q) = \kappa(\vx_p, \vx_q) = \exp\left(-\frac{\|\vx_p - \vx_q\|_2^2}{2\sigma^2}\right)
	\\
    &= \exp\left(-\frac{\sum_{j=1}^J(\vx_p^{(j)} - \vx_q^{(j)})^2}{2\sigma^2}\right) \\
    &= \exp\left(\frac{-(\vx_p^{(1)} - \vx_q^{(1)})^2}{2\sigma^2}\right) 
	\cdots\exp\left(\frac{-(\vx_p^{(J)} - \vx_q^{(J)})^2}{2\sigma^2}\right) \\
    &= \vK^{(1)}(p,q) \cdot \vK^{(2)}(p,q) \cdot \cdots \cdot \vK^{(J)}(p,q),
\end{aligned}
\end{equation}
where $\sigma$ is the kernel width. 
Using $\circ$ to denote the Hadamard (element-wise) product operator, we rewrite (\ref{equ: element RBF}) as $\vK = \vK^{(1)} \circ \vK^{(2)} \circ \cdots \circ \vK^{(J)}.$
Therefore, once the eigenvectors of each local kernel matrix are obtained, the whole kernel matrix $\vK$ could be approximated as the following way,
\begin{equation}\label{equ: agg-rbf}
\begin{aligned}
	\hat{\vK} &=\hat{\vK}^{(1)} \circ \cdots \circ \hat{\vK}^{(J)}\\
    &=  \left(\vV_D^{(1)}\bm{\lambda}_D^{(1)}(\vV_D^{(1)})^\top\right) \circ \cdots \circ  \left(\vV_D^{(J)}\bm{\lambda}_D^{(J)}(\vV_D^{(J)})^\top\right).
\end{aligned}
\end{equation}
Finally, we compute the first $D$ eigenvectors of $\hat{\vK}$, denoted as $\bm \hat \vV$, and the projection matrix $\hat{\vW} = \sum_i \hat{\vV}_i \phi(\vx_i)$. Notice that for this calculation, $\vW$ is unknown but $\langle\vW,\phi(\vy)\rangle$ can be calculated by kernel trick in a distributed system because $\vx^\top \vy$ and $\|\vx-\vy\|_F^2$ is separable along features.
The overall algorithm is summarized in Algorithm~\ref{alg: DKPCA-os}.

\textbf{Approximation Analysis.}
Here we present the approximation analysis for DKPCA in both linear and RBF cases.
Specifically, we study  
the $\sin\Theta$ distance between the eigenspaces spanned by $\vV$ and $\hat{\vV}$, where ${\vV}$ are the eigenvectors of the global gram kernel matrix $\vK$, $\hat{\vV}$ is the estimator calculated by DKPCA.
$\sin\Theta$ distance is well-defined and is widely used for measuring the distance between two linear spaces \cite{fan2019distributed, yu2015useful}.  
Let $\alpha_1,\cdots, \alpha_D$ be the singular values of $\vV^\top\hat{\vV}$ and define $\sin\Theta(\vV, \hat{\vV})$ as follows.
\begin{equation}\label{equ: sin distance}
\begin{aligned}
	\Theta(\vV, \hat{\vV}) &= \mathrm{diag}\{\cos^{-1}(\alpha_1),\cdots, \cos^{-1}(\alpha_d)\}\\
	&\triangleq\mathrm{diag}\{\theta_1,\cdots, \theta_d\} \\
	\sin\Theta(\vV, \hat{\vV}) 
	&= \mathrm{diag}\{\sin(\theta_1),\cdots, \sin(\theta_d))\}.
\end{aligned}	
\end{equation}

\begin{theorem}\label{the: linearspaces bound theorem}
Let $\vV \in \mathcal{R}^{T\times D}$ be the first $D$ eigenvectors of the global kernel matrix $\vK \in \mathcal{R}^{T\times T}$ that is derived by a kernel function $\kappa$, and $\bm \hat{\vV}$ be its approximation computed by DKPCA.  
If $\kappa$ is a linear kernel, then  $\bm{\vV,\hat{\vV}}$ satisfy
\begin{equation}\label{equ: linear bound}
	\|\sin\Theta(\vV, \hat{\vV})\|_F \leq \frac{J\sqrt{T-D}\max_j(\lambda^{(j)}_{D+1})}{\lambda_D(\vK)-\lambda_{D+1}(\vK)}.
\end{equation}
If $\kappa$ is a RBF kernel, then  $\bm{\vV,\hat{\vV}}$ satisfy
\begin{equation}\label{equ: RBF bound}
    	\|\sin\Theta(\vV, \hat{\vV})\|_F \leq \frac{J\sqrt{T}\max_j(\lambda^{(j)}_{D+1})}{\lambda_D(\vK)-\lambda_{D+1}(\vK)}.
\end{equation}
\end{theorem}
See supplemental materials for the proof of Theorem~\ref{the: linearspaces bound theorem}.
Theorem~\ref{the: linearspaces bound theorem} indicates that the approximation error is related with $J, T,D$ and $\lambda$.
If $\lambda$ decay fast, which is common for RBF kernels, then DKPCA will have very high quality results.

\textbf{Communication and Computation cost.}
\begin{table}
    \centering
    \caption{Communication and computation complexity of three kinds of algorithms for PCA.}
    \label{tab: cost}
    \begin{tabular}{|c|c|c|}
    \hline
         & COMM. & COMP.   \\ \hline
        DPCA \cite{fan2019distributed} & $\mathcal{O}(DM)$ & $\mathcal{O}(\max\{M, (T/J + DJ)\}M^2)$   \\ \hline
        DKPCA (Alg.~\ref{alg: DKPCA-os}) & $\mathcal{O}(DT)$ & $\mathcal{O}(\max\{T, (M/J + DJ)\}T^2)$  \\ \hline
        KPCA (SVD-based) & $\mathcal{O}(TM/J)$  & $\mathcal{O}(T^3)$ \\
        \hline
    \end{tabular}
\end{table}
Alg.~\ref{alg: DKPCA-os} is quite efficient with only one round communication. To analyze quantitatively, we restrict our discussion on the evenly distributed situation, i.e., the local feature dimension is $\mathcal{O}(M/J)$ and there is no statistic difference among the nodes. The discussion on other cases is similar but is more complicated in form. 
We summarize the communication and computation complexity of distributed PCA (DPCA) \cite{fan2019distributed}, Alg.~\ref{alg: DKPCA-os} and central KPCA in Table~\ref{tab: cost}.

Compared with central algorithms, where additional communication and fusion are required, DKPCA sacrifices computation efficiency for communication efficiency.
To pursue high communication efficiency, we prefer a small $D$, e.g., when $D$ is much smaller than $M/J$, DKPCA has significant advantage over central algorithms on communication cost.
As for the computation cost, for given data, if $T \geq 2\sqrt{M(D+1)} $ and
\begin{equation*}
    \begin{aligned}
        J\in  & \left[\frac{T-\sqrt{\Delta}}{2(D+1)}, \frac{T+\sqrt{\Delta}}{2(D+1)}\right] \mathrm{~with~}  \Delta=T^2-4M(D+1).
    \end{aligned}
\end{equation*}
Then the computation cost is $\mathcal{O}(T^3)$, the same as central algorithms. 
Notice that the required condition is not strict.
For example, if $M=10000$, $D=100$, $T=5000$, then $J\in [3, 47]$, which is a large range, will meet the above requirement such that the computation cost is $\mathcal{O}(T^3)$.

\textbf{Self-adaptive strategy for data maldistribution.}
Hereinbefore, we simply set equal $D^{(j)}$ in every local agents for DKPCA, which, however, may not work well for the case of data maldistribution.
Thm.~\ref{the: linearspaces bound theorem} shows that for given $J,T,D$\footnote{In fact, $J,T,\delta$ are the inherent attribute of data that we can not change. We will show the different performance of DKPCA for data with different $J,T,\delta$ in section \ref{sec: exp}.}, the error is related to $\delta$ and $\mathrm{max}_j \lambda^{(j)}_{D+1}$.
To reduce the approximate error, small $\lambda^{(j)}_{D+1}$ is preferred. Thus, local agents that have larger eigenvalues (which means they hold more information) need to send more number of eigenvectors.
On the other side, in some agents, the eigenvalues decay fast so that the $D$-th one are almost $0$. Then sending $D$ eigenvectors is redundant. 

Therefore, we consider the following self-adaptive strategy for the selection of $D^{(j)}$ such that each agent can decide the number of the eigenvectors sent to the center according to their local eigenvalues $\bm \lambda^{(j)}$:
\begin{equation}
\label{equ: adaptive }
D^{(j)} = \min\{\min_D\{\lambda_{D+1}(\vK^{(j)})\leq {\epsilon}\}, T\},
\end{equation}
where $\epsilon$ is a positive threshold value.
Note that the decays of $\bm \lambda_i$ are generally assumed to be polynomial or exponential \cite{bach2013sharp}, i.e. $\lambda_i = \mathcal{O}(i^{-\delta})$ or  $\lambda_i = \mathcal{O}(e^{-\rho i})$. Then the $D$ given by (\ref{equ: adaptive }) is  $\Omega((\frac{c}{\epsilon})^{\frac{1}{\delta}})$ or $\Omega(-\frac{1}{\rho}\log(\frac{\epsilon}{c}))$, where $c$ is some constant.

\section{Experiments}\label{sec: exp}
\begin{figure}[t]
    \centering
    \subfloat[]{\includegraphics[width=0.24\textwidth]{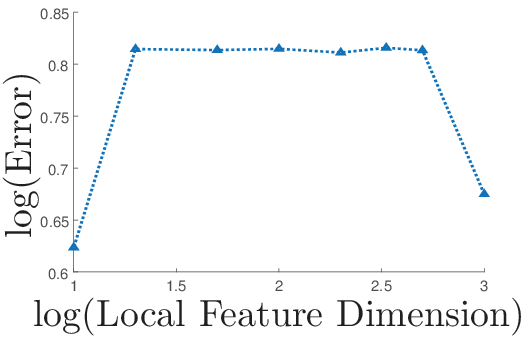}}
    \subfloat[]{\includegraphics[width=0.24\textwidth]{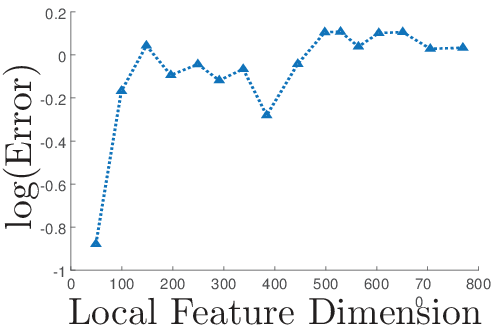}}
    \caption{The log of the mean error with respect to the number of local agents. (a) Simulation data and linear kernels with $M=1000,D=10$ and $T=400$. (b) Real data and RBF kernels ($\sigma = \frac{\sqrt{M}}{3}$) with $M=8545,D=10$ and $T=537$.}
    \label{fig: expA}
\end{figure}

\begin{figure*}[tbp]
    \centering
    \subfloat[]{\includegraphics[width=0.24\textwidth]{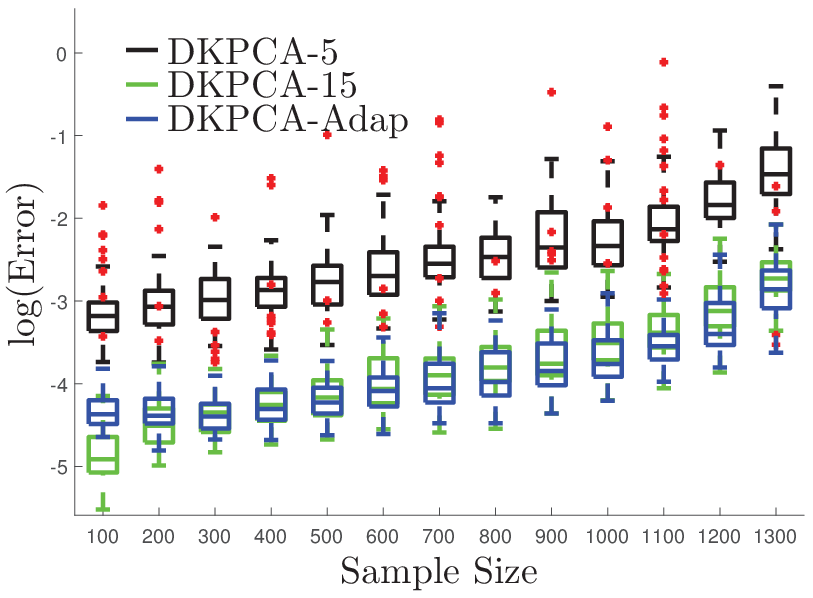}}
    \subfloat[]{\includegraphics[width=0.24\textwidth]{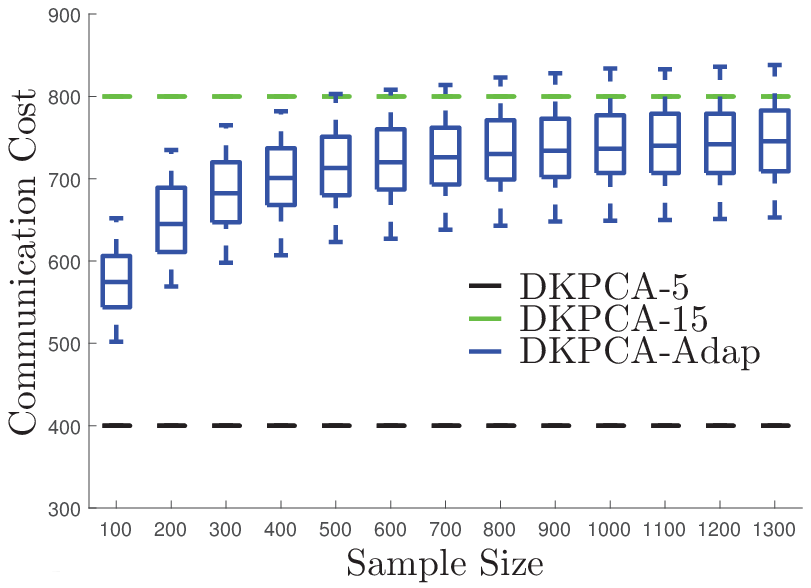}}
    \subfloat[]{\includegraphics[width=0.24\textwidth]{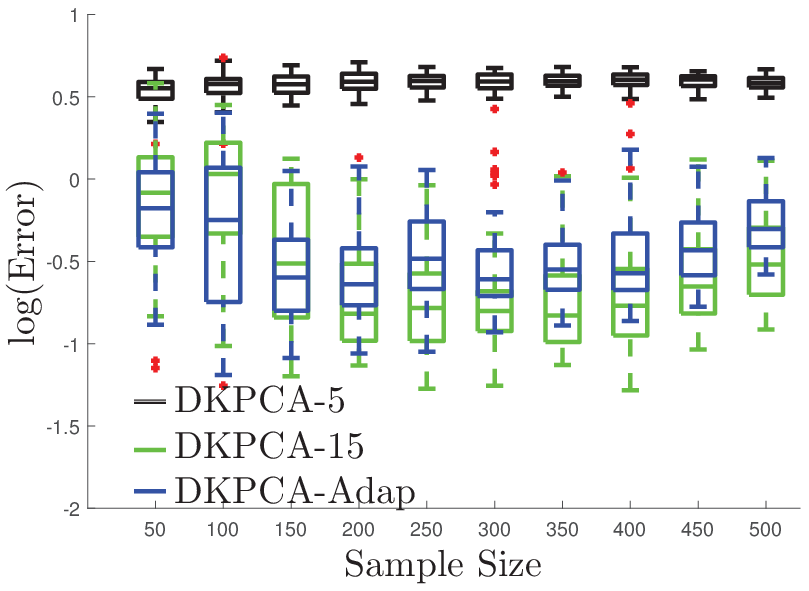}}
    \subfloat[]{\includegraphics[width=0.24\textwidth]{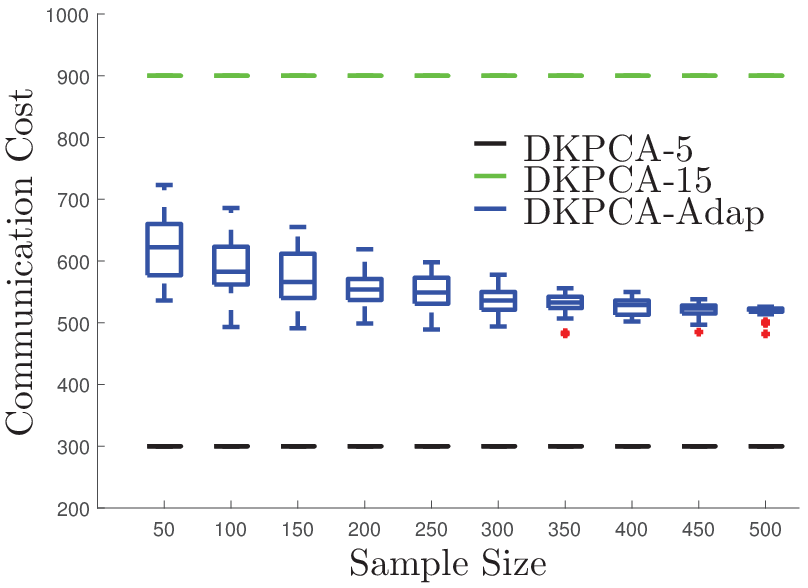}}
    \caption{The log of the mean error and the total number of local eigenvectors that transmitted to the center with respect to the number of samples. (a-b) Simulation data and the linear kernel are used, where  $M=1000,\;D=10$ and $J=10$. (c-d) Real data and the RBF kernel are used  with $\sigma = \frac{\sqrt{M}}{3}$, $M=8545,\;D=10$ and $J=60$.}
    \label{fig: expB}
\end{figure*}

\begin{table*}[tbp]
\scriptsize
\centering 
\caption{Classify error rate on GSE2187 dataset using L-SVM with DKPCA, KPCA and PCA.}
\label{tab: GSE}
\begin{tabular}{c|c|c|c|c|c|c|c|c}
\toprule
\multicolumn{2}{c|}{\begin{tabular}[c]{@{}c@{}}Number \\ of Features\end{tabular}} & 1 & 5 & 10 & 20 & 50 & 100 & 200 \\ \midrule
\multirow{3}{*}{\begin{tabular}[c]{@{}c@{}}Toxicant\\vs\\ Fibrate\end{tabular}}               
& DKPCA & 0.3675$\pm$0.0383   & 0.0323$\pm$0.0218  & 0.0164$\pm$0.0128   & 0.0150$\pm$0.0118  & 0.0148$\pm$0.0111   & 0.0145$\pm$0.0110     & 0.0148$\pm$0.0115    \\ \cline{2-9} 
& KPCA          & 0.3675$\pm$0.0383  & 0.0259$\pm$0.0151  & 0.0173$\pm$0.0124   & 0.0148$\pm$0.0111  & 0.0150$\pm$0.0111   &  0.0145$\pm$0.0110  &  0.0148$\pm$0.0115   \\ \cline{2-9} 
& PCA           & 0.5048$\pm$0.0932  & 0.2009$\pm$0.1121  & 0.0352$\pm$0.0517   & 0.0259$\pm$0.0551  & 0.0277$\pm$0.0702   &  0.0264$\pm$0.0587  &  0.0243$\pm$0.0562   \\ \hline
\multirow{3}{*}{\begin{tabular}[c]{@{}c@{}}Toxicant \\vs\\ Azole\end{tabular}}               
& DKPCA & 0.4790$\pm$0.0410  & 0.3966$\pm$0.0305  & 0.2778$\pm$0.0361   & 0.1907$\pm$0.0376  & 0.1416$\pm$0.0331   & 0.1099$\pm$0.0283    &  0.0969$\pm$0.0259   \\ \cline{2-9} 
& KPCA          & 0.4889$\pm$0.0350  & 0.3912$\pm$0.0341  & 0.2778$\pm$0.0379   & 0.1844$\pm$0.0341  & 0.1391$\pm$0.0325   &  0.1102$\pm$0.0283  &  0.0969$\pm$0.0259   \\ \cline{2-9} 
& PCA           & 0.5019$\pm$0.0616  & 0.5124$\pm$0.0415  & 0.4655$\pm$0.0620   & 0.3781$\pm$0.0782  & 0.2207$\pm$0.1309   & 0.1982$\pm$0.1502  &  0.2134$\pm$0.1671   \\ \hline
\multirow{3}{*}{\begin{tabular}[c]{@{}c@{}}Toxicant \\vs\\ Others\end{tabular}}               
& DKPCA & 0.3341$\pm$0.0148  & 0.3326$\pm$0.0202  & 0.2477$\pm$0.0209   & 0.2004$\pm$0.0231  & 0.1499$\pm$0.0197   &  0.1428$\pm$0.0169   &  0.1410$\pm$0.0172   \\ \cline{2-9} 
& KPCA          & 0.3341$\pm$0.0148  & 0.3284$\pm$0.0267  & 0.2420$\pm$0.0229   & 0.1973$\pm$0.0207  & 0.1488$\pm$0.0190   &  0.1430$\pm$0.0176  &  0.1410$\pm$0.0172   \\ \cline{2-9} 
& PCA           & 0.4578$\pm$0.1367  & 0.4928$\pm$0.0992  & 0.4129$\pm$0.0763   & 0.3526$\pm$0.0888  & 0.2045$\pm$0.1324   & 0.1953$\pm$0.1613  &  0.2077$\pm$0.1584   \\ \bottomrule
\end{tabular}
\end{table*}

The performance of DKPCA is evaluated here by three experiments, where both
simulation and real data are used for linear cases and non-linear cases, respectively. 
Details of simulation data generation is in supplemental materials.

Real data used in our experiments include the gene expression of different drugs and toxicants on rats \cite{natsoulis2005classification}, which is collected on cRNA microarray chips and is available at the NIH GEO, under accession number GSE2187.
The total data is $8565\times 537$, corresponding to four categories: fibrates (107 samples), statins (93 samples), azoles (156 samples) and toxicants (181 samples). 
The features are removed if more than $10\%$ of the samples have their values missing. The rest missing values are filled with mean values. 

The first $D$ eigenvectors $\vV_{gt} \in \mathcal{R}^{D\times T}$ calculated by performing the SVD algorithm on the whole underlying kernel matrix are regarded as the ground truth. 
We use $\sin \Theta$ distance to evaluate the error of the estimator $\hat{\vV} \in \mathcal{R}^{D\times T}$, i.e., $\mathrm{Error}= D - \|\vV_{gt}^\top\hat{\vV}\|_F^2.$
All the simulations are repeated for 50 times and are done with Matlab R2016b in Core i5-7300HQ 2.50GHz 8GB RAM. The codes of DKPCA and experiments are available in \url{https://github.com/hefansjtu/DKPCA.}

\textbf{Relationship between Estimation Error and the Number of Local Agents.}
In this experiment, we vary the number of local agents to see how it affects the estimation error.
The result is reported in Fig.~\ref{fig: expA}, where (a) linear kernels with $M=1000$ and $T=400$ and (b) RBF kernels with $\sigma = \frac{\sqrt{M}}{3},M=8545$ and $T=537$ are considered. 
The data are uniformly partitioned and the target is to estimate the first $D=10$ eigenvectors of the global kernel matrix $\vK$.

Fig~\ref{fig: expA} shows that overall the estimation error is small.
For different numbers of local agents, the estimation error is similar except the extreme cases: there is only one agent or each agent transmits all the data.
This phenomenon can be explained by Theorem~\ref{the: linearspaces bound theorem}, which indicates that the error bound has a positive correlation to the number of local agents $J$ and the $D$-th local eigenvalues.
It should be noted that the information in local decreases when $J$ increases, which further leads to the decay of local eigenvalues.
Thus, even when the number of local agents increases, the accuracy could be maintained at a high level.

\textbf{The Self-adaptive Strategy for Data Maldistribution.}
The error of DKPCA and DKPCA with self-adaptive strategy (DKPCA-Adap) are reported in Fig.~\ref{fig: expB} with linear and RBF kernels.
The target is to recover the first $D=10$ eigenvectors in data maldistribution cases, where the local feature dimensions are randomly generated.
The result of DKPCA with fixed $D^{(j)}=C$ is denoted as DKPCA-$C$.
$\epsilon$ in DKPCA-Adap is set as $0.04\lambda_1$ for linear cases and $0.0005\lambda_1$ for non-linear cases, where $\lambda_1$ denotes the largest eigenvalue in local.
Different $T$ are also considered to see how the accuracy changes.

Fig.~\ref{fig: expB} demonstrates the relationship between the sample size and (a) the estimation error (b) the total communication cost $\sum_j{D^{(j)}}$. 
Intuitively, the performance of DKPCA-$5$ is worst because it requires the least communication and thus DKPCA-$15$ is the best. 
The DKPCA-Adap achieves similar accuracy as the DKPCA-$15$ but requires much less communication, showing the effectiveness of the self-adaptive strategy.

It is interesting that the tendency of communication cost with respect to $T$ are different in Fig.~\ref{fig: expB},
which increases in linear cases but decreases in RBF cases.
The main reason is the rank of the data.
The simulation data is low-rank and thus adding new sample brings little new information.
Therefore, the increase of sample size forces all the eigenvalues grow uniformly, leading to the increase of $D^{(j)}$ and thus the high communication cost.
But the high-dimension real data is small-size, adding new data causes the increase of principal eigenvalue (more than other eigenvalues), leading to the decrease of the communication cost.
In conclusion, this strategy is more suitable for small distributed dataset with high dimension.

\textbf{Comparison between distributed and full sample KPCA in real classification tasks.}
The aim of PCA is to keep useful information during data projection. 
Here we will show that DKPCA can preserve similar information as KPCA.
That is, the same post learning algorithm can achieve similar performance on the two projected data produced by DKPCA and KPCA.
We first map data into a low-dimension feature space by DKPCA, the central kernel algorithm (KPCA) with RBF kernels with $\sigma = 50$, and the central linear algorithm (PCA), for which the corresponding eigenproblems are all solved by SVD.
The feature dimension of the projected data changes from $1$ to $200$.
The projected data are then sent a linear support vector machine (L-SVM). We randomly choose $200$ data as the training set and use the rest for test. 

The average classification error and its standard deviation over 50 trials are reported in Table~\ref{tab: GSE}. 
As the feature dimension of the projected data increases, the classification error rates of all methods decrease. 
KPCA is based on full data and is expected to be better than the proposed DKPCA. But from Table~\ref{tab: GSE}, one could observe that the difference is slight. In KPCA and DKPCA, the RBF kernel is applied and thus achieves less classification error than PCA, although the PCA is conducted on full data. Notice that before the proposed DKPCA, there is no distributed PCA algorithm that can apply nonlinear kernels in vertical regime.

\section{Conclusion}
This letter introduced a one-shot privacy-preserving algorithm, DKPCA, for distributed kernel PCA in vertical regime. 
The main technique is kernel trick, by which the vertical regime and horizontal regime is linked up and nonlinear dimension reduction could be implemented with RBF kernels.
We presented the approximation analysis and experiments result of DKPCA, which coincides with the theoretical result and demonstrate that DKPCA is effective to extract non-linear features and is communication efficient.

\newpage

\bibliographystyle{IEEEtran}
\bibliography{thesis.bib}

\end{document}